\documentclass[anonymous]{article}
\usepackage{ijcai25}

\usepackage{times}
\usepackage{soul}
\usepackage{url}
\usepackage[hidelinks]{hyperref}
\hypersetup{breaklinks=true}
\usepackage[utf8]{inputenc}
\usepackage[small]{caption}
\usepackage{graphicx}
\usepackage{amsthm}
\usepackage{booktabs}
\usepackage{algorithm}
\usepackage{algorithmic}
\usepackage{amsmath} 
\usepackage{amssymb} 
\usepackage{multirow}
\usepackage[switch]{lineno}

\title{Conditional Denoising Meets Polynomial Modeling: A Flexible Decoupled Framework for Time Series Forecasting}

\author{
Jintao Zhang\textsuperscript{\rm 1},
Mingyue Cheng\textsuperscript{\rm 1}\footnote{Corresponding author},
Xiaoyu Tao\textsuperscript{\rm 1},
Zhiding Liu\textsuperscript{\rm 1},
Daoyu Wang\textsuperscript{\rm 1}
\affiliations
\textsuperscript{\rm 1}State Key Laboratory of Cognitive Intelligence, University of Science and Technology of China \\
\emails
zjttt@mail.ustc.edu.cn, mycheng@ustc.edu.cn, \{txytiny, zhiding, wdy030428\}@mail.ustc.edu.cn
}

\begin{document}

\maketitle

\begin{abstract}
Time series forecasting models are becoming increasingly prevalent due to their critical role in decision-making across various domains. However, most existing approaches represent the coupled temporal patterns, often neglecting the distinction between their specific components. In particular, fluctuating patterns and smooth trends within time series exhibit distinct characteristics. In this work, to model complicated temporal patterns, we propose a Conditional Denoising Polynomial Modeling (CDPM) framework, where probabilistic diffusion models and deterministic linear models are trained end-to-end. Instead of modeling the coupled time series, CDPM decomposes it into trend and seasonal components for modeling them separately. To capture the fluctuating seasonal component, we employ a probabilistic diffusion model based on statistical properties from the historical window. For the smooth trend component, a module is proposed to enhance linear models by incorporating historical dependencies, thereby preserving underlying trends and mitigating noise distortion. Extensive experiments conducted on six benchmarks demonstrate the effectiveness of our framework, highlighting the potential of combining probabilistic and deterministic models. Our code is available at \url{https://github.com/zjt-gpu/CDPM}.
\end{abstract}

\section{Introduction}

Time series forecasting is a fundamental task across various domains, playing a crucial role in driving informed decision-making in industries such as finance, healthcare, and manufacturing~\cite{intro1}. It involves analyzing historical patterns to model the dynamic evolution of underlying processes and predict future trends~\cite{cheng2025instructime}. Accurate time series forecasting is essential for tasks such as demand prediction, resource allocation, enabling organizations to anticipate future needs and optimize operations~\cite{intro6,liu2024generative}.

\begin{figure}[hbpt]
    \centering
    \includegraphics[width=\linewidth]{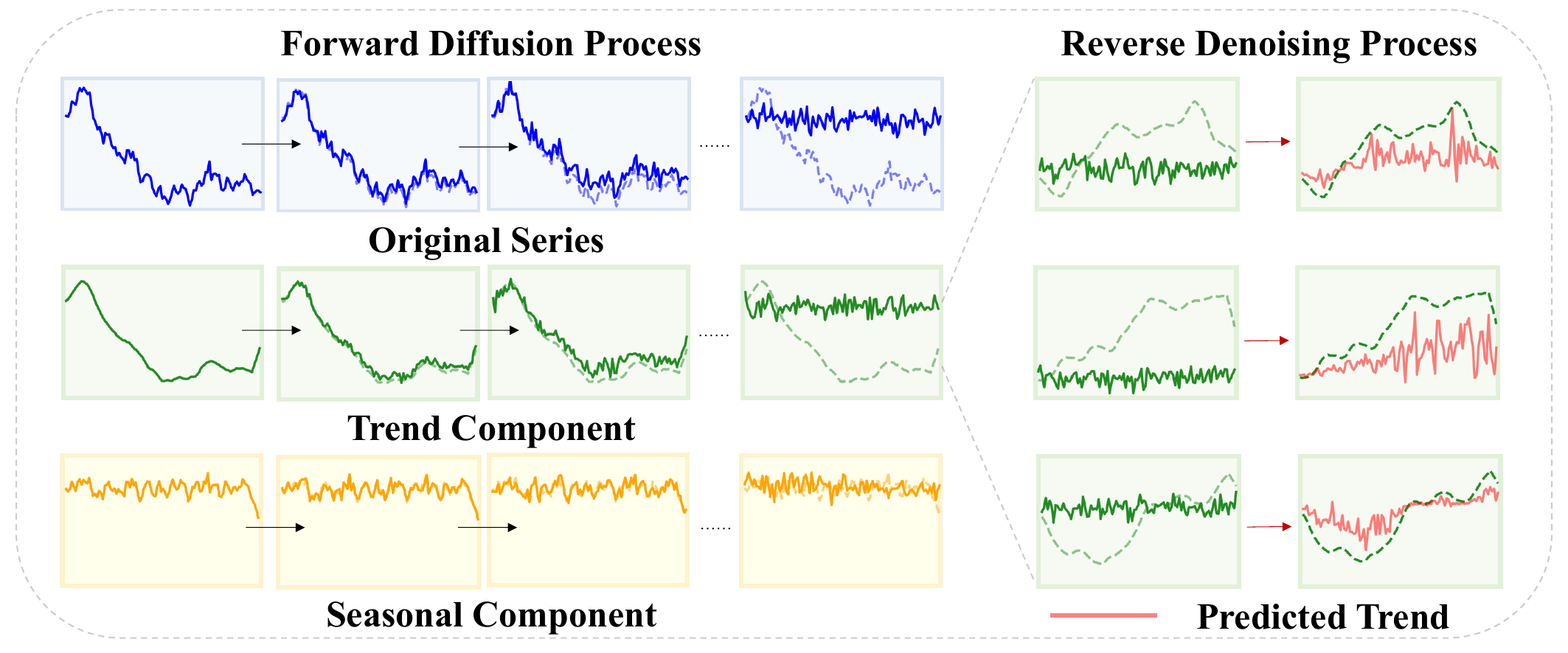}
    \caption{The forward diffusion process iteratively adds noise to the original time series, which causes the trend components to be overshadowed as the noise increasingly dominates.}
    \label{fig:motivation}
\end{figure}

Over the years, various approaches have been proposed to address this challenge~\cite{cheng2025comprehensive}, ranging from classical statistical models, such as Autoregressive Integrated Moving Average (ARIMA)~\cite{ARIMA}, to machine learning techniques~\cite{ML}, and advanced deep learning models, including recurrent neural networks (RNNs)~\cite{SegRNN,LSTNet}, convolutional neural networks (CNNs)~\cite{TCN,CNNijcai}, and transformer-based models~\cite{Informer,Autoformer}. Large language models (LLMs) such as TimeLLM and LLMTIME~\cite{Time-llm,LLMTIME} have demonstrated the potential in time series forecasting. In addition to their success, probabilistic models have emerged as promising alternatives, offering a principled framework for quantifying uncertainty and modeling complex distributions. Diffusion models~\cite{Timegrad,TimeDiff}, becoming the leading paradigm for probabilistic models originally developed for generation, have proven effective in time series forecasting due to their ability to capture intricate stochastic processes. 

Diffusion-based methods~\cite{lee2024ant} have shown great promise in time series forecasting. However, we argue that their naive application of noise addition and denoising overlooks the heterogeneous nature of temporal dynamics. In particular, these methods uniformly add noise across the entire sequence and apply denoising without distinguishing between different temporal components. However, time series inherently exhibits diverse dynamics, including trends and seasonal patterns, which require different treatment. Trends patterns, reflecting gradual shifts over extended periods, and seasonal patterns, characterized by periodic fluctuations, possess different structural properties. These distinct patterns require differentiated treatment. Such indiscriminate processing can distort essential features, especially intrinsic trends and fluctuations, thereby degrading forecasting performance. As shown in Fig.\ref{fig:motivation}, this indiscriminate approach disrupts critical trend information during denoising, hindering trend recovery and reducing accuracy.

Typically, a single model often struggles to capture the subtle and dynamic interplay between trends and variations, which can significantly affect the model's overall forecasting ability. Diffusion models, which excel at modeling complex dynamic processes, are more suitable for handling components characterized by large fluctuations and high-frequency variations in time series. In contrast, deterministic models, such as models based on linear layers, learn features from historical patterns and are particularly effective at identifying regular patterns, making them better suited for modeling components with stable trends. Therefore, by strategically leveraging the strengths of both model types and tailoring them to the distinct characteristics of time series, we can more fully capture the dynamic variations and enhance the final forecasting results.

To overcome these challenges, we propose a novel Conditional Denoising Polynomial Modeling (CDPM) framework that leverages the inductive bias of time series decomposition. By separating the time series into trend and seasonal components, we enable decoupled modeling tailored to the characteristics of each component. Specifically, we employ a carefully designed Conditional Denoising Seasonal Module (CDSM) to capture the high-frequency seasonal component, introducing an innovative denoising strategy conditioned on statistical properties derived from historical window. Currently, we model the intrinsic trend component using the polynomial trend module (PTM), which is adept at capturing gradual changes. Additionally, we reformulate the Improved Evidence Lower Bound (ELBO) to facilitate joint training of the CDSM and PTM. This decoupling method of our framework improves the ability to capture temporal dependencies and complex patterns. The key contributions of this work are:

\begin{itemize}
    \item We introduce a novel framework that enables the decoupled modeling of trend and seasonal components, providing new insights into time series forecasting.
    \item We design tailored methods and reformulate ELBO for each component-using CDSM for the seasonal component and PTM for the trend component—to effectively capture their distinct behaviors.
    \item We conduct comprehensive experiments on benchmark datasets to validate effectiveness of our framework, we also present the rationale for modeling different components with several case studies.
\end{itemize}

\begin{figure*}
    \centering
    \includegraphics[width=1\linewidth]{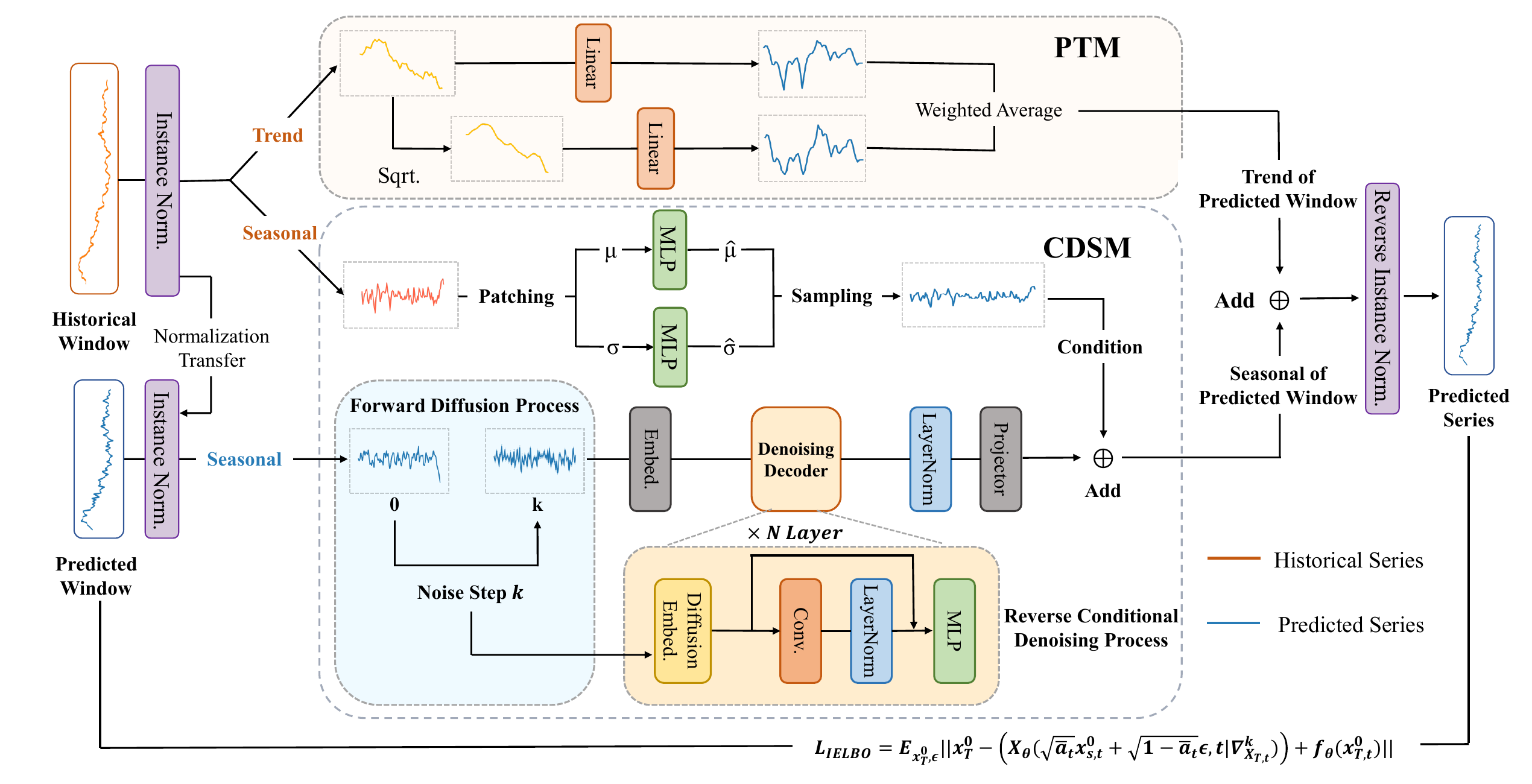}
    \caption{Illustration of the proposed Conditional Denoising Polynomial Modeling (CDPM) framework. The Conditional Denoising Seasonal Module (CDSM) refines the seasonal component through a diffusion process conditioned on historical patterns, while the Polynomial Trend Module (PTM) models the trend component to capture different patterns.}
    \label{fig:network}
\end{figure*}

\section{Related Work}

\subsection{Deterministic Forecasting}

Deterministic models map deterministic inputs to deterministic outputs by learning consistent patterns within time series, making them crucial for capturing predictable temporal relationships. Recurrent neural networks (RNNs), such as LSTNet~\cite{LSTNet} and SegRNN~\cite{SegRNN}, became popular for sequential dependencies, while convolutional neural networks (CNNs)~\cite{cheng2024convtimenet} like TCN~\cite{TCN} and SCINet~\cite{SCINet} excelled at capturing both local and global temporal features. Graph-based models, such as MTGNN~\cite{MTGNN} and FourierGNN~\cite{FourierGNN}, improved forecasting by modeling interdependencies between time series. More recently, Transformer-based models like Informer~\cite{Informer} and Autoformer~\cite{Autoformer} have utilized self-attention mechanisms to efficiently capture long-range dependencies, excelling in high-dimensional and long-horizon tasks. Surprisingly, well-tuned linear models such as DLinear~\cite{DLinear} and TSMixer~\cite{Tsmixer} have demonstrated strong performance by effectively capturing long-term trends, often outperforming more complex architectures and showcasing their efficiency. Despite advances, deterministic models often struggle with volatile fluctuations, limiting their ability to model complex time series dynamics.

\subsection{Diffusion-Based Time Series Forecasting}
Probabilistic models offer uncertainty estimation and the ability to model temporal distributions. These models, including Variational Autoencoders (VAEs)\cite{VAE}, Generative Adversarial Networks (GANs)\cite{GAN}, and flow-based models~\cite{flow}, are designed to capture high-dimensional distribution. Recently, Denoising Diffusion Probabilistic Models (DDPMs) have emerged as a promising paradigm for generative modeling, demonstrating their potential in various time series applications. DDPMs have advanced time series forecasting by capturing complex temporal patterns. Early models like TimeGrad~\cite{Timegrad} introduced autoregressive denoising with Langevin sampling for multivariate predictions. TSDiff~\cite{TSDiff} improved short-term accuracy through self-guiding mechanisms, while score-based models such as ScoreGrad~\cite{ScoreGrad} applied stochastic differential equations (SDEs) for continuous-time forecasting, expanding diffusion to fluctuatingly sampled data. Conditional models like TimeDiff~\cite{TimeDiff} incorporated external information to guide the diffusion process, outperforming traditional methods. Latent Diffusion Models (LDMs)\cite{imageLDT} enhanced efficiency by performing diffusion in lower-dimensional spaces. This approach was further demonstrated by LDCast and Latent Diffusion Transformers (LDT)\cite{LDT}, which improved precipitation and time series forecasting. Models like DSPD and CSPD~\cite{DSPD} applied diffusion to anomaly detection and interpolation. 

Despite the advancements in both deterministic and probabilistic forecasting approaches, existing research typically applies each paradigm separately to represent time series, with limited efforts to integrate them into a unified modeling framework. However, effectively capturing the diverse characteristics of time series, such as the smooth trend and fluctuating seasonal components, requires leveraging both deterministic and probabilistic models. 

\section{Problem Statement}

Time series represent numerical sequence ordered chronologically, typically recorded at regular temporal intervals. Formally, consider a multivariate time series \( X_{L+T} \in \mathbb{R}^{(L+T) \times d} \), where \( L \) and \( T \)  denotes the length of the historical and the length of the predicted window respectively, with \( d \) channels recorded for each time step. The objective of time series forecasting is to train a model \( G \) that learns to map the historical series \( X_{L} \) to the predicted series \( X_{T} \). The model leverages the information embedded in the historical series to predict future series, aiming to capture both temporal dependencies for accurate forecasting.

\section{Methodology}

In this section, we formalize the time series forecasting problem and present the proposed Conditional Denoising Polynomial Modeling (CDPM) framework, comprising two core components: the Conditional Denoising Seasonal Module (CDSM) and the Polynomial Trend Module (PTM).

\subsection{Overall Framework}

The CDPM addresses time series forecasting by independently modeling trend and seasonal patterns, thereby capturing distinct temporal behaviors more effectively. As depicted in Fig.~\ref{fig:network}, CDPM consists of two main modules: CDSM and PTM. The CDSM employs a denoising diffusion process conditioned on statistical properties from the historical window to accurately capture and refine seasonal variation patterns. In parallel, the PTM models trends using polynomial linear layers, providing flexibility in handling both linear and nonlinear trends. By decoupling and specifically modeling these components, CDPM enhances forecasting accuracy, offering an effective solution for capturing the complex temporal dynamics inherent in time series. We also reformulate the Evidence Lower Bound (ELBO) to better jointly train the PTM and CDSM. By handling trend and seasonal components independently, CDPM allows each module to focus on its respective dynamics, leading to enhanced overall performance.

\subsection{Temporal Feature Structuring}

\paragraph{Instance Normalization.}

Instance normalization across both historical and predicted windows is essential for maintaining the stability of the CDPM. We normalize the historical window by subtracting the mean and dividing by the standard deviation~\cite{kim2021reversible}, and transfer the same statistics to normalize the predicted window during training. Specifically, \( x_L \) is normalized to \( x_L^0 \), and \( x_T \) is normalized to \( x_T^0 \) using the statistics derived from \( x_L \). This ensures alignment between the windows and preserves both the temporal structure and stability in the forecasting process. The subsequent denormalization step restores the predicted values to their original scale, ensuring consistency with the input data distribution.

\paragraph{Decomposition Operation.}

Decomposition is an important approach to achieving decoupling in time series analysis, as it allows for the separation of trend components and seasonal components~\cite{STL}, enabling each characteristic to be handled independently. We decompose both historical \( X_L^0 \) and predicted \( X_T^0 \) windows into trend and seasonal components. The trend component \( X_{L,t}^0 \) of historical window is estimated using a moving average:

\begin{align}
X_{L,t}^0 = \text{Avg}(X_L^0),
\end{align}
where \( \text{Avg} \) denotes the moving average.

\begin{align}
X_{L,s}^0 = X_L^0 - X_{L,t}^0,\label{historical}
\end{align}
where the seasonal component \( X_{L,s}^0 \) is the residual after subtracting the trend. Thus, we obtain the trend component \( X_{L,t}^0 \) and seasonal component \( X_{L,s}^0 \) of historical window.

Similarly, we compute \( X_{T,t}^0 \) and \( X_{T,s}^0 \). Decomposition operation isolates the distinct characteristics of time series.

\subsection{Conditional Denoising Seasonal Module}

Modeling fluctuating seasonal components in time series is challenging due to their intricate patterns. We propose CDSM, based on a denoising diffusion model. By incorporating historical statistical properties as conditional inputs, CDSM captures seasonal patterns more effectively.

\paragraph{Forward Diffusion Process.}

In the forward diffusion process, noise is progressively added to the seasonal component. At each diffusion step \( k \), the noisy seasonal series is:

\begin{align}
X_{T,s}^k = \sqrt{\bar{\alpha}_k} \, x_{T,s}^0 + \sqrt{1 - \bar{\alpha}_k} \, \epsilon,
\end{align}
where \( X_{T,s}^k \) is the seasonal component at step \( k \), \( x_{T,s}^0 \) is the original seasonal component, \( \epsilon \sim \mathcal{N}(0, I) \) is Gaussian noise, and \( \bar{\alpha}_k \in [0,1] \) is the noise schedule. As \( k \) increases, the seasonal component becomes more noise-dominated.

\paragraph{Reverse Conditional Denoising Process.}

The denoising process starts by embedding \( X_{T,s}^k \):

\begin{align}
\mathbf{h} = \text{Emb}(X_{T,s}^k),\label{predicted}
\end{align}
where \( \text{Emb}(\cdot) \) is a convolutional MLP that maps the noisy sequence to a hidden dimension while capturing local temporal features. Sinusoidal positional encodings for diffusion step \( t \) are integrated via Adaptive Layer Normalization (AdaLN):

\begin{align}
\text{AdaLN}(\mathbf{h}) = \alpha_t \cdot \text{LayerNorm}(\mathbf{h}) + \beta_t,
\label{add_noise}
\end{align}
where \( \alpha_t \) and \( \beta_t \) are learnable parameters projected from the diffusion step encoding.

The decoder composed of convolutional layers and MLPs, captures global dependencies and feature interactions. Convolutional layers extract fine-grained local temporal variations, while MLPs model cross-dimensional relationships. Layer normalization and a projection layer map the output back to the original feature space.

To enhance denoising process, we incorporate conditional information from the historical seasonal component \( X_{L,s}^0 \).

\begin{equation}
\mu_{L,i} = \frac{1}{H} \sum_{h=1}^{H} x_{L,s,i,h}, 
\end{equation}

\begin{equation}
\sigma_{L,i}^2 = \frac{1}{H} \sum_{h=1}^{H} \left( x_{L,s,i,h} - \mu_{L,i} \right)^2,
\end{equation}
where, for each patch \( P_{L,i} \), \( \mu_{L,i} \) represents the local mean and \( \sigma_{L,i}^2 \) denotes the variance. The historical data is divided into patches \( P \) of length \( H \).

These statistics capture seasonal variability. The predicted mean \( \hat{\mu}_{T,i} \) and variance \( \hat{\sigma}_{T,i}^2 \) for the target window \( P_{T,i} \) are computed using two MLPs:

\begin{equation}
\hat{\mu}_{T,i} = \text{MLP}_{\mu}(\mu_{L,i}), 
\end{equation}

\begin{equation} 
\hat{\sigma}_{T,i}^2 = \text{MLP}_{\sigma}(\sigma_{L,i}^2),
\end{equation}
where \(\text{MLP}_{\mu}(\cdot)\) and \(\text{MLP}_{\sigma}(\cdot)\) capture the mean and variance of the historical window features, respectively. These predicted statistics are subsequently used in the denoising process via Gaussian sampling:

\begin{equation}
\nabla_{\mathbf{x}_{T,s}}^{k-1} = \hat{\mu}_{T,i} + \hat{\sigma}_{T,i} \cdot z,\label{conditional}
\end{equation}
where \( z \sim \mathcal{N}(0, I) \). This enables the model to incorporate stochastic variability while aligning with historical patterns.

Finally, the denoised output \( \hat{x}_{T,s}^0 \) is combined with the conditional information through a weighted summation:

\begin{align}
X_\theta \left( X_{T,s}^k, t \mid \nabla_{\mathbf{x}_{T,s}}^{k-1} \right) = \rho_1 \hat{x}_{T,s}^0 + \rho_2 \nabla_{\mathbf{x}_{T,s}}^{k-1},\label{season}
\end{align}
where \( \rho_1 \) and \( \rho_2 \) are learnable parameters that balance the denoised series and conditional information, ensuring effective seasonal modeling while maintaining statistical properties.

\subsection{Polynomial Trend Module}

Different trends follow distinct trajectories, including phases of growth and decline, highlighting the need to capture diverse trend behaviors. By leveraging Cover's theorem~\cite{covergeometrical}, we project smooth time series patterns into a polynomial space, facilitating the extraction of both linear and nonlinear trends. Based on this, polynomial modeling~\cite{Polynomial} is employed to better capture these diverse trend variations. To fully harness polynomial modeling, we introduce a square root transformation to smooth series and enhance the framework’s flexibility in capturing diverse trend patterns.

As a deterministic model, PTM consists of two complementary pathways. The first models stable trends with a linear layer applied directly to the historical trend series:

\begin{align}
T_{\text{origin}}(x_{T,t}^0) = \text{Linear}_{\text{origin}}(x_{T,t}^0),
\end{align}
where \(\text{Linear}_{\text{origin}}(\cdot)\) captures the general trend.

The second pathway enhances trend modeling by applying square root transformation, thereby mitigating the impact of extreme values and stabilizing trend predictions:

\begin{align}
T_{\text{sqrt}}(x_{T,t}^0) = \text{Linear}_{\text{root}}(\sqrt{x_{T,t}^0}),
\end{align}
where \(\text{Linear}_{\text{sqrt}}(\cdot)\) captures the transformed trend.

The outputs from both pathways are combined via weighted summation to capture a wider range of trends:

\begin{align}
f_\theta(x_{T,t}^0) = \lambda_1 T_{\text{origin}}(x_{T,t}^0) + \lambda_2 T_{\text{sqrt}}(x_{T,t}^0),\label{trend}
\end{align}
where \(\lambda_1\) and \(\lambda_2\) are learnable parameters that balance the contributions of the linear and root-based components. This mechanism allows the model to adapt to both linear and sub-linear growth patterns, enhancing its ability to forecast complex time series with diverse dynamics.

\subsection{Optimization}
Integrating trend and seasonal components improves forecasting accuracy by combining their strengths for more comprehensive predictions. Finally, the derived loss function enables joint optimization, allowing both components to be learned concurrently and effectively.

\paragraph{Component Reconstruction.}

The final prediction \( \hat{x}_T^0 \) is calculated by integrating the trend \( \hat{x}_{T,t}^0 \) and seasonal components \( \hat{x}_{T,s}^0 \) as follows:

\begin{align}
\hat{x}_T^0 = X_\theta \left( \sqrt{\bar{\alpha}_t} x_{s,t}^0 + \sqrt{1 - \bar{\alpha}_t} \epsilon, t \mid \nabla_{\mathbf{x}_{T,s}}^{k} \right) + f_\theta(x_{T,t}^0),\label{Integration}
\end{align}
where \( X_\theta(\cdot) \) represents the CDSM and \( f_\theta(\cdot) \) corresponds to the PTM, respectively. 

\paragraph{Improved ELBO Optimization.}

The final prediction \( \hat{x}_T^0 \) is subsequently denormalized to \( \hat{x}_T \) using the mean and variance derived from the historical window, ensuring that the forecast is consistent with the original distribution of the time series. The loss function in CDPM is derived from the Evidence Lower Bound (ELBO). Furthermore, we extend the traditional ELBO to accommodate the separate modeling of the trend and seasonal components. This decoupled modeling of the trend and seasonal components ensures that both contribute independently to the final forecast. Through derivation, we find that by modeling the seasonal component with CDSM and the trend component with PTM, the results can be jointly optimized and trained to improve the overall performance of our framework, leading to more accurate predictions. The improved ELBO is given by:

\begin{align}
L_{\text{IELBO}} = \mathbb{E}_{x_T^0, \epsilon} \left\| x_T^0 - \hat{x}_T^0 \right\|^2,\label{loss}
\end{align}
where \( x_T^0 \) and \( \hat{x}_T^0 \) represent the true and predicted values of the time series, respectively. This formulation combines the strengths of both probabilistic diffusion models and deterministic linear models, leading to more accurate forecasts.

\begin{table}[h]
  \centering
  \resizebox{0.47\textwidth}{!}{
  \begin{tabular}{ccccccc}
    \toprule
    Dataset & ETTh1 & ETTm1 & Wind & Exchange & Weather & Electricity \\
    \midrule
    Dimension       & 7   & 7   & 7   & 8   & 21  & 321  \\
    Frequency & 1 hour & 15 mins & 15 mins & 1 day & 10 mins & 1 hour \\
    T (Steps) & 168 & 192 & 192 & 14  & 672 & 168  \\
    \bottomrule
  \end{tabular}}
  \caption{Overview of the dataset characteristics.}
  \label{tab:dataset_statistics}
\end{table}

\section{Experiments}
\subsection{Experimental Settings}

\begin{table*}
  \centering
  \footnotesize
  \renewcommand{\arraystretch}{1.2}
  \resizebox{\textwidth}{!}{
    \begin{tabular}{cc|cc|cccccccccccccccc}
    \toprule
    \multirow{2}[0]{*}{Dataset} & \multirow{2}[0]{*}{T} & \multicolumn{2}{c|}{CDPM} & \multicolumn{2}{c}{iTransformer} & \multicolumn{2}{c}{Patchmixer} & \multicolumn{2}{c}{TSMixer} & \multicolumn{2}{c}{PatchTST} & \multicolumn{2}{c}{DLinear} & \multicolumn{2}{c}{Diffusion-TS} & \multicolumn{2}{c}{D3VAE} & \multicolumn{2}{c}{CSDI} \\
          &       & MSE   & MAE   & MSE   & MAE   & MSE   & MAE   & MSE   & MAE   & MSE   & MAE   & MSE   & MAE   & MSE  & MAE   & MSE   & MAE   & MSE   & MAE \\
    \midrule
     ETTh1 & 168   & \textbf{0.4364} & \textbf{0.4363} & 0.4433 & 0.4429 & 0.4610 & 0.4560 & 0.4418 & \underline{0.4401} & 0.4401 & 0.4463 & \underline{0.4372} & 0.4420 & 1.6559 & 1.0400 & 0.9523 & 0.8618 & 1.1090 & 0.8009  \\
     ETTh2 & 168 & \textbf{0.3393} & \textbf{0.3826} & 0.4009 & 0.4156 & 0.3731 & 0.4055 & \underline{0.3654} & \underline{0.3981} & 0.3805 & 0.4036 & 0.3976 & 0.4233 & 3.2225 & 1.4736 & 5.3165 & 1.9796 & 2.0700 & 1.0748 \\
    ETTm1 & 192   & \textbf{0.3532} & \textbf{0.3781} & 0.3685 & 0.3916 & 0.3731 & 0.3983 & 0.3673 & 0.3924 & \underline{0.3626} & \underline{0.3890} & 0.3706 & 0.3968 & 1.6769 & 1.0326 & 0.6341 & 0.6833 & 1.0770 & 0.7788 \\
    ETTm2 & 192   & \textbf{0.2303} & \textbf{0.3009} & 0.2535 & 0.3184 & 0.2643 & 0.3226 & 0.2495 & \underline{0.3108} & \underline{0.2472} & 0.3129 & 0.2498 & 0.3224 & 2.9380 & 1.3732 & 4.1001 & 1.5884 & 1.6051 & 0.9397 \\
    Electricity & 168   & 0.1606 & 0.2526 & \textbf{0.1531} & \textbf{0.2461} & 0.1559 & 0.2505 & 0.1563 & 0.2497 & \underline{0.1537} & \underline{0.2494} & 0.1628 & 0.2581 & 1.4888 & 0.9935 & 1.7061 & 0.9273 & —— & —— \\
    Exchange & 14    & \textbf{0.0204} & \underline{0.0979} & 0.0719 & 0.1896 & 0.0430 & 0.1150 & \underline{0.0209} & \textbf{0.0955} & 0.0293 & 0.1178 & 0.0412 & 0.1320 & 2.7192 & 1.3573 & 4.8311 & 1.9942 & 1.3023 & 0.8719 \\
    Weather & 672   & \textbf{0.3292} & \textbf{0.3373} & 0.3676 & 0.3709 & \underline{0.3295} & 0.3411 & 0.3316 & 0.3430 & 0.3299 & \underline{0.3400} & 0.3398 & 0.3801 & —— & —— & —— & —— & 0.4842 & 0.4498 \\
    Wind  & 192   & 1.1262 & 0.7534 & \textbf{1.0653} & \textbf{0.7227} & 1.1497 & 0.7582 & 1.2407 & 0.7845 & 1.1411 & 0.7607 & \underline{1.0826} & \underline{0.7378} & 2.0706 & 1.1185 & 3.6837 & 1.4195 & 1.4913 & 0.8942 \\
    \bottomrule
    \end{tabular}%
  }
  \caption{The forecasting results on eight datasets, with the best results highlighted in bold and the second-best underlined. CSDI encounters out-of-memory issues on the Electricity and Diffusion-TS datasets, while D3VAE runs out of memory on the Weather dataset.}
  \label{tab:overall}%
\end{table*}

\paragraph{Datasets.}

We evaluate the effectiveness of our proposed framework on multiple real-world time series datasets from various domains. The datasets include ETTh and ETTm~\cite{Informer}, which represent electricity transformer temperature data; Exchange, which contains exchange rates from eight countries; and Weather\footnote{\url{https://www.bgc-jena.mpg.de/wetter/}}, consisting of 21 meteorological indicators sampled at 10-minute intervals. Additionally, we use the Electricity dataset\footnote{\href{https://archive.ics.uci.edu/ml/datasets/ElectricityLoadDiagrams20112014}%
{https://archive.ics.uci.edu/ml/datasets/ElectricityLoadDiagrams\linebreak20112014}}, which records electricity consumption from 321 clients, and Wind~\cite{D3VAE}, comprising wind power measurements sampled every 15 minutes from 2020 to 2021.

To ensure consistency in chronological data splitting, we adopt a 6:2:2 split for the ETTh and ETTm datasets, and a 7:1:2 split for the Wind, Weather, Electricity, and Exchange datasets. This setup enables comprehensive testing of the framework's generalization across different temporal segments. Consistent with TimeDiff~\cite{TimeDiff}, we evaluate performance using historical window lengths of ${96, 192, 720, 1440}$ to cover varying forecasting horizons.

\paragraph{Baselines.}

We compare our proposed Conditional Denoising Polynomial Modeling (CDPM) framework against a diverse set of baseline models.
\begin{itemize}
    \item \textbf{Diffusion models:} CSDI~\cite{CSDI}, Diffusion-TS~\cite{Diffusion-TS}.
    \item \textbf{Transformer-based models:} iTransformer~\cite{iTransformer}, PatchTST~\cite{PatchTST}.
    \item \textbf{Linear-based models:} DLinear~\cite{DLinear}, TSMixer~\cite{Tsmixer}. 
    \item \textbf{Hybrid model:} PatchMixer~\cite{Patchmixer}, D3VAE~\cite{D3VAE}. 
\end{itemize}

\paragraph{Implementation Details.}

Our proposed CDPM is implemented using PyTorch and trained with Exponential Learning Rate Scheduler~\cite{ExponentialLR}, starting with an initial learning rate of $1 \times 10^{-3}$ and a batch size of 16. Early stopping with a patience of 10 epochs is employed to prevent overfitting. The diffusion process utilizes $K = 50$ steps with a cosine variance schedule~\cite{Timegrad} from $\beta_1 = 10^{-4}$ to $\beta_K = 0.5$. We adopt the DDIM sampler~\cite{DDIM}. All experiments are conducted on a workstation equipped with an NVIDIA GeForce RTX 4090 GPU.

\subsection{Experiment Results}

\paragraph{Main Results.}

The experimental results in Table~\ref{tab:overall} validate the effectiveness of the CDPM for time series forecasting. CDPM’s decoupling strategy, which separates trend and seasonal components, allows for more accurate modeling of temporal patterns. It performs well on challenging datasets like ETTh1 and ETTh2, where traditional models such as Diffusion-TS and D3VAE struggle with varying time resolutions and high noise. CDPM's strength lies in its Polynomial Trend Module (PTM) and Conditional Denoising Seasonal Module (CDSM), which capture complex trends and seasonal dynamics. Unlike CSDI and Diffusion-TS, which struggle with long forecast horizons, CDPM handles long-term dependencies effectively. On volatile datasets like Wind, CDPM captures sharp fluctuations and achieves competitive results, though further refinement could improve performance on highly noisy data. Overall, CDPM shows strong generalization across various datasets.

\paragraph{Ablation Study.}

\begin{table}[h!]
  \centering
    \small
    \begin{tabular}{lcccc}
    \toprule
    \multirow{2}{*} {Method} & \multicolumn{2}{c}{ETTh2} & \multicolumn{2}{c}{ETTm2} \\
    & MSE   & MAE  & MSE   & MAE \\ 
    \midrule
    CDPM       & \textbf{0.3393} & \textbf{0.3826} & \textbf{0.2303} & \textbf{0.3009} \\
    \text{CDPM\textsuperscript{1}} & 0.3404 & 0.3781 & 0.2320 & 0.3023 \\
    \text{CDPM\textsuperscript{2}} & 0.3433 & 0.3826 & 0.2450 & 0.3121 \\
    \text{CDPM\textsuperscript{3}} & 0.3853 & 0.4215 & 0.2459 & 0.3272 \\
    \text{CDPM\textsuperscript{4}} & 0.3442 & 0.3807 & 0.2335 & 0.3015 \\
    \bottomrule
    \end{tabular}%
    \caption{Ablation study results on ETTh2 and ETTm2.}
  \label{tab:ablation_results}%
\end{table}

The ablation experiments highlight the importance of each component in the CDPM framework, especially for handling complex temporal dynamics in datasets with varying granularities like ETTh2 and ETTm2. We present the results of ablation experiments in Table~\ref{tab:ablation_results} to evaluate the contributions of individual components within the CDPM framework.

\begin{itemize}
    \item \textbf{\text{CDPM\textsuperscript{1}}}: Removes the conditional information.
    \item \textbf{\text{CDPM\textsuperscript{2}}}: Replaces the PTM with a simple linear layer.
    \item \textbf{\text{CDPM\textsuperscript{3}}}: Removes instance normalization alignment.
    \item \textbf{\text{CDPM\textsuperscript{4}}}: Removes the conditional information and replaces PTM with a linear layer.
\end{itemize}

Removing the condition information injection module reduces performance, with a larger impact on ETTm2, where external factors influence intricate temporal patterns. Replacing the PTM with a linear layer causes a significant performance drop, which is crucial for capturing long-term trends and periodic behaviors. Eliminating instance normalization leads to the largest performance decline, highlighting its role in stabilizing training for high-resolution datasets.  Removing both the condition information injection module and PTM further degrades performance, emphasizing the necessity of both components.

\begin{table}[h!]
  \centering
    \small
    \begin{tabular}{lcccc}
    \toprule
    \multirow{2}{*} {Method} & \multicolumn{2}{c}{ETTh2} & \multicolumn{2}{c}{ETTm2} \\
    & MSE   & MAE   & MSE   & MAE \\ 
    \midrule
    CDPM      & \textbf{0.3393} & \textbf{0.3826} & \textbf{0.2303} & \textbf{0.3009} \\
    \text{CDPM\textsuperscript{5}} & 0.3839 & 0.4106 & 0.2518 & 0.3157 \\
    \text{CDPM\textsuperscript{6}} & 0.3541 & 0.3832 & 0.2452 & 0.3122 \\
    \bottomrule
    \end{tabular}%
    \caption{Decoupling validation study results on ETTh2 and ETTm2.}
  \label{tab:ablation_Decoupling}%
\end{table}

\paragraph{Evaluation of the Decoupling Strategy.}

The Decoupling Validation Study emphasizes the importance of using appropriate models to represent the distinct components of the time series, while also providing evidence for the effectiveness and validity of the decoupling strategy. We present the results of ablation experiments in Table~\ref{tab:ablation_Decoupling} to evaluate the validation of our decoupling framework.

\begin{itemize}
    \item \textbf{\text{CDPM\textsuperscript{5}}}: This variant directly applies the CDSM to the original time series.
    \item \textbf{\text{CDPM\textsuperscript{6}}}: In this variant, PTM and CDSM are reversed, with PTM predicting the seasonal component and CDSM predicting the trend component.
\end{itemize}

The decoupling strategy that separates trend and seasonal patterns markedly improves performance.  Applying our framework directly underperforms due to its inability to separate trend and seasonal components, while reversing PTM and CDSM shows slight improvements but still falls short of our framework, highlighting the importance of proper module assignment. These results emphasize the need to tailor modules for distinct patterns, with decoupling especially beneficial for datasets with clear temporal structure.

\begin{figure*}
    \centering
    \includegraphics[width=1\linewidth]{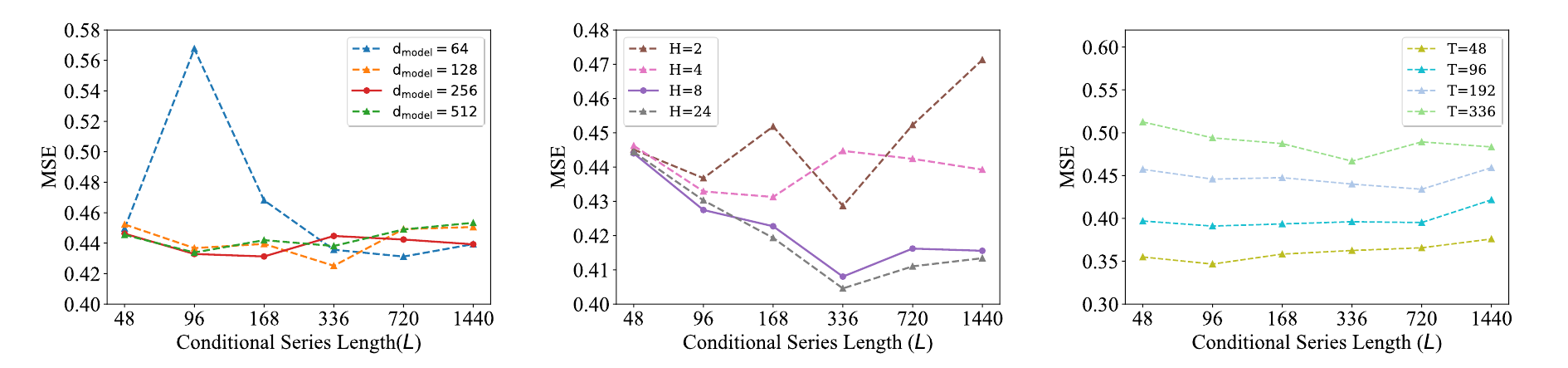}
    \caption{Parameter sensitivity analysis of CDPM: hidden dimension (left), patch length (middle), and predicted length (right).}
    \label{fig:hyperparam_analysis}
\end{figure*}

\begin{figure*}[htbp]
    \centering
    \includegraphics[width=0.9\linewidth]{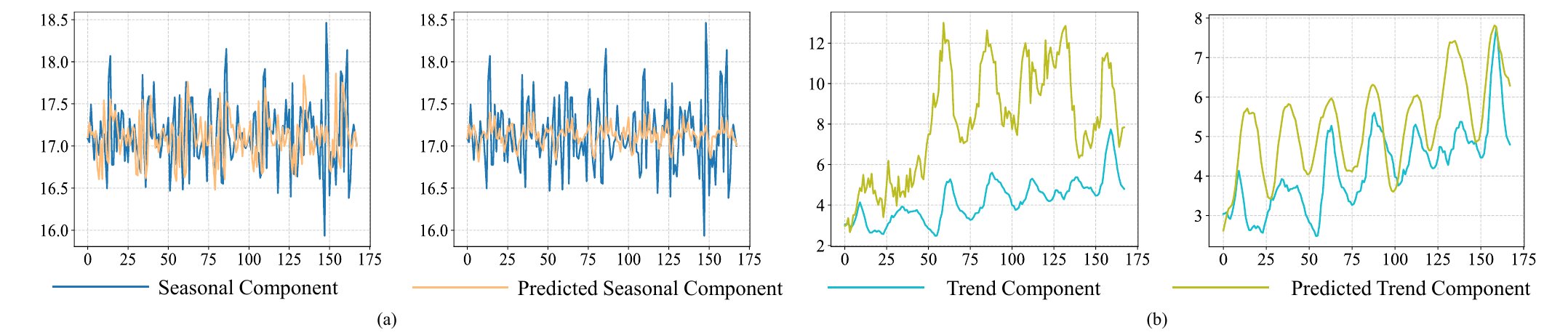}  
    \caption{(a) Visual comparison of CDSM (left) and PTM (right) in modeling seasonal components. (b) Visual comparison of CDSM (left) and PTM (right) in modeling trend components.}
    \label{fig:trend&seasonal_comparison}
\end{figure*}

\paragraph{PTM and CDSM Modeling Analysis.}

To evaluate the effectiveness of PTM and CDSM in modeling different components of the ETTh1 dataset, we conducted experiments with an input length of 192 and a prediction horizon of 168. The results highlight the effectiveness of tailored modeling strategies for each component. As shown in Figure~\ref{fig:trend&seasonal_comparison} (a), PTM excels in capturing intrinsic and gradual trends due to its linear structure, accurately modeling patterns and tracking the overall direction of the time series. In contrast, applying CDSM to the trend component introduces unnecessary noise and overfitting to minor variations, diminishing trend quality. For the seasonal component, which exhibits complex fluctuations, the seasonal module demonstrates superior performance, as shown in Figure~\ref{fig:trend&seasonal_comparison} (b). These results validate each module's modeling capabilities for its designated component.

\paragraph{Hyper-Parameter Sensitivity Analysis.}

To investigate the impact of key design choices in CDPM, such as hidden dimension size ($d_{\text{model}}$), patch length ($H$), and prediction horizon on model performance, we conducted a hyperparameter sensitivity analysis using the ETTh1 dataset. The results show that increasing $d_{\text{model}}$ initially improves performance, peaking at $d_{\text{model}} = 256$, while larger values (e.g., $d_{\text{model}} = 512$) offer diminishing returns and risk overfitting, highlighting the trade-off between model capacity and efficiency. Similarly, $H$ significantly affects performance, with $H = 8$ capturing both local dependencies and global patterns, whereas larger values introduce redundancy and reduce efficiency. Finally, our framework performs consistently across prediction horizons.

\begin{figure}
    \centering
    \includegraphics[width=1\linewidth]{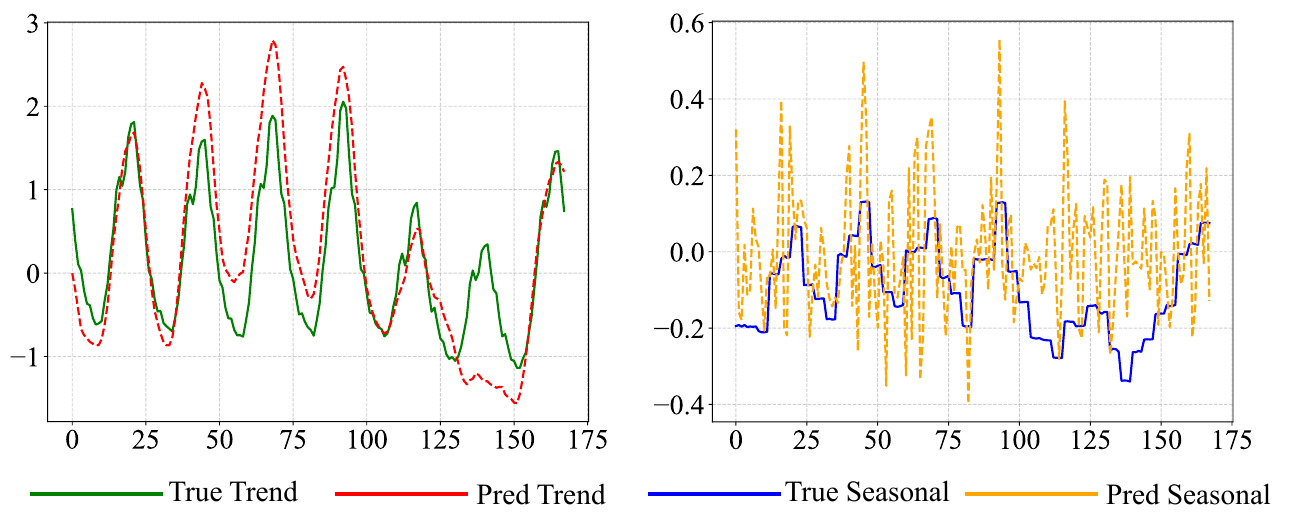}
    \caption{Visualization of trend (left) and seasonall (right) components of joint modeling for the electricity dataset.}
    \label{fig:electricity_1}
\end{figure}

\paragraph{Combined Modeling Capability Analysis.}

To assess the interpretability and predictive performance of the framework, we conducted a joint analysis of the PTM and CDSM modules on the Electricity dataset, as shown in Figure~\ref{fig:electricity_1}. The framework decouples multidimensional time series, capturing diverse trends and cyclic patterns, demonstrating its capacity to model complex temporal dynamics. In Figure~\ref{fig:electricity_1} (left), applied to the Electricity dataset with clear periodic and trend structures, the trend module captures the underlying patterns, with predictions closely matching actual values. Figure~\ref{fig:electricity_1} (right) illustrates the seasonal module's ability to model periodic variations, where predicted components align well with the observed sequences, compensating for deviations in trend estimation and improving accuracy. The integration of deterministic and diffusion-based modules within the framework enhances both interpretability and predictive accuracy across datasets with combined components.

\section{Conclusion}

In this paper, we proposed the Conditional Denoising Polynomial Modeling (CDPM), a novel time series forecasting framework that separately models trend and seasonal components using specialized modules, trained in an end-to-end manner. The Polynomial Trend Module (PTM) captured intrinsic trends, while the Conditional Denoising Seasonal Module (CDSM) addressed complex seasonal fluctuations. By reformulating the Evidence Lower Bound, we were able to better integrate the training of both modules. This decoupling paradigm enhanced the modeling of temporal dynamics, resulting in improved forecasting accuracy. Extensive experiments on real-world datasets validated the effectiveness of the proposed CDPM.

\clearpage

\section*{Acknowledgements}

This research was supported by grants from the grants of Provincial Natural Science Foundation of Anhui Province (No.2408085QF193), the National Natural Science Foundation of China (62337001), the Key Technologies R\&D Program of Anhui Province (No. 202423k09020039), USTC Research Funds of the Double First-Class Initiative (No. YD2150002501), and the Fundamental Research Funds for the Central Universities (No. WK2150110032).

\bibliographystyle{named}
\bibliography{ijcai25}

\end{document}